\documentclass[10pt,twocolumn,letterpaper]{article}

\usepackage{cvpr}
\usepackage{times}
\usepackage{epsfig}
\usepackage{graphicx}
\usepackage{amsmath}
\usepackage{amssymb}
\usepackage{paralist} 

\usepackage[T1]{fontenc}
\usepackage[utf8]{inputenc}
\usepackage{authblk}
\usepackage{multirow}
\usepackage{makecell}
\usepackage{adjustbox}
\usepackage{algorithm2e}
\usepackage{verbatim}

\usepackage[pagebackref=true,breaklinks=true,letterpaper=true,colorlinks,bookmarks=false]{hyperref}

\cvprfinalcopy 

\def\cvprPaperID{5183} 

\DeclareMathOperator*{\argmax}{arg\,max}
\ifcvprfinal\fi
\begin{document}

\title{FGN: Fully Guided Network for Few-Shot Instance Segmentation} 
\author{{Zhibo Fan$^{1}$},
        {Jin-Gang Yu$^{1,2,}\thanks{Corresponding author}$}, \ 
        {Zhihao Liang$^{1}$}, 
        {Jiarong Ou$^{1}$}, \\ \vspace{-8pt}
        {Changxin Gao$^{3}$},
        {Gui-Song Xia$^{4}$},
        {Yuanqing Li$^{1,2}$}
        
        \vspace{4pt}
        $^{1}$South China University of Technology \ \  
       $^{2}$Guangzhou Laboratory
       
       $^{3}$Huazhong University of Science and Technology \ \
       $^{4}$Wuhan University
       
       \vspace{4pt}
       \small{\{zanefan0323,zhliang19980922\}@gmail.com, \{jingangyu,yqli\}@scut.edu.cn,
      
      au\_jaring@mail.scut.edu.cn, cgao@hust.edu.cn, guisong.xia@whu.edu.cn}}

\maketitle
\begin{abstract}

Few-shot instance segmentation (FSIS) conjoins the few-shot learning paradigm with general instance segmentation, which provides a possible way of tackling instance segmentation in the lack of abundant labeled data for training. This paper presents a Fully Guided Network (FGN) for few-shot instance segmentation. FGN perceives FSIS as a guided model where a so-called support set is encoded and utilized to guide the predictions of a base instance segmentation network (\textit{i.e.}, Mask R-CNN), critical to which is the guidance mechanism. In this view, FGN introduces different guidance mechanisms into the various key components in Mask R-CNN, including Attention-Guided RPN, Relation-Guided Detector, and Attention-Guided FCN, in order to make full use of the guidance effect from the support set and adapt better to the inter-class generalization. Experiments on public datasets demonstrate that our proposed FGN can outperform the state-of-the-art methods.   
\end{abstract}

\vspace{-18pt}
\section{Introduction}
\thispagestyle{empty}

Instance segmentation~\cite{hariharan2014simultaneous, he2017mask} is a fundamental computer vision task which aims to simultaneously localize, classify and  estimate the segmentation masks of object instances from a given image. The past few years have witnessed notable advances on instance segmentation thanks to the prosperity of convolutional neural networks (CNN)~\cite{he2017mask, liu2018path, chen2018masklab, chen2019hybrid}, as well as its success in a variety of real-world applications~\cite{zhang2016instance, yu2019exemplar,gupta2014learning}. Existing CNN-based approaches to instance segmentation are mostly fully-supervised, which require abundant labeled data for model training~\cite{he2017mask,ren2017end,hayder2017boundary}. Such a data-hungry setting however may be impractical.

\begin{figure}[!h]
\centering
    \includegraphics[width = 8cm]{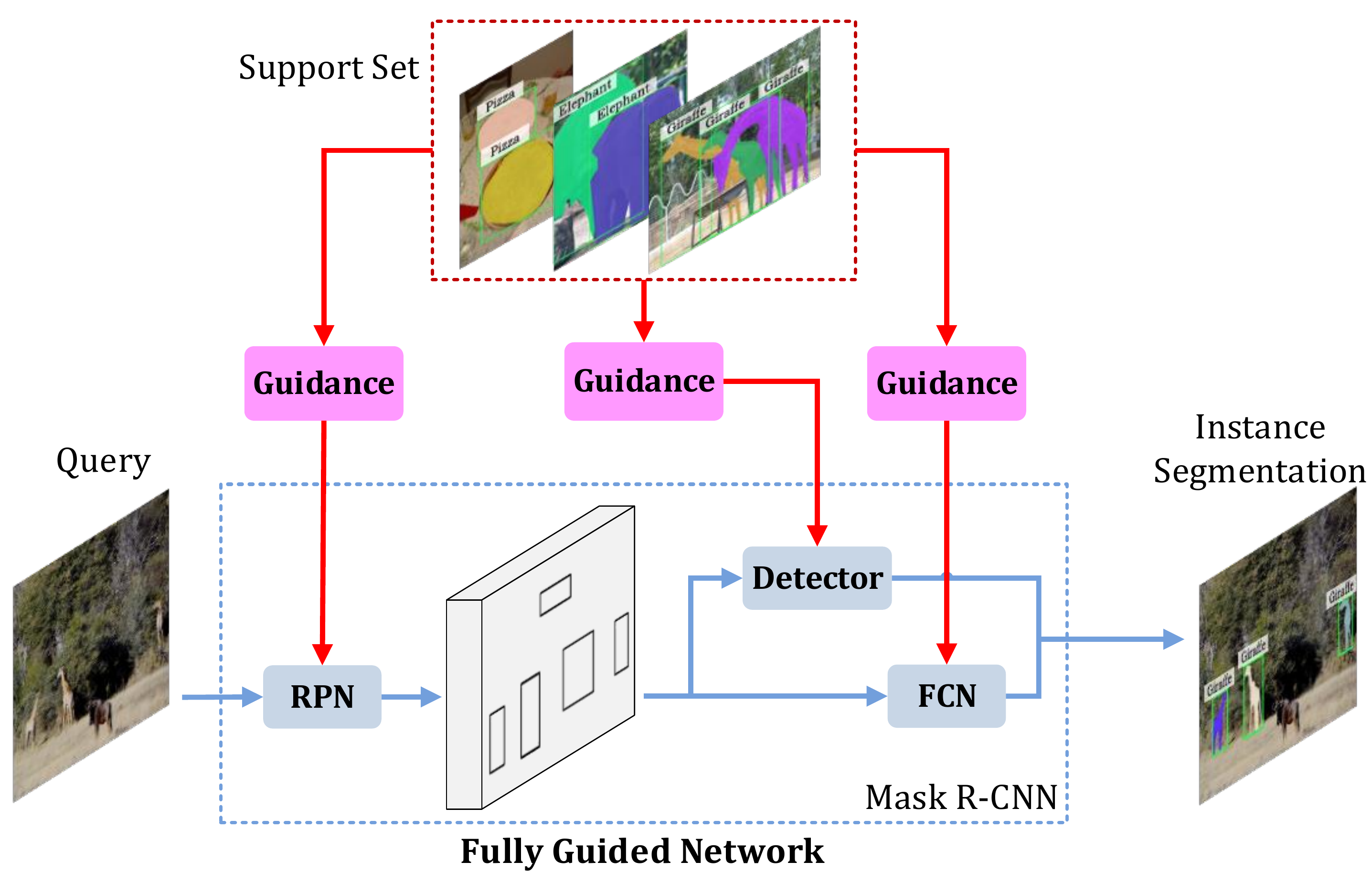}
	\caption{Illustration of few-shot instance segmentation using the proposed Fully Guided Network (FGN). To adapt better to the inter-class generalization, FGN introduces different guidance mechanisms for the various key components in Mask R-CNN. }
\label{fig:fsis-concept}
\vspace{-1.5\baselineskip}
\end{figure}


Inspired by the remarkable ability of human to learn with limited data, few-shot learning (FSL) has recently received a lot of research attention~\cite{vinyals2016matching,snell2017prototypical,koch2015siamese,sung2018learning,garcia2018few}. Assuming the availability of a large amount of labeled data belonging to certain classes (base classes) for training, FSL aims at making predictions on data from other different classes (novel classes) given only a handful of labeled exemplars for each~\cite{vinyals2016matching, snell2017prototypical}. Instead of fine-tuning an ordinary model pre-trained on base classes with the very limited novel-class samples, or conducting data augmentation, FSL learns a conditional model that makes predictions conditioned on a support set, so as to adapt to the inter-class generalization.

The majority of existing FSL models focus on visual classification, and a minority on semantic segmentation~\cite{zhang2018sg,rakelly2018conditional,shaban2017one,dong2018few}. Nevertheless, it has been rarely explored so far in the context of instance segmentation, the task of our concern termed as few-shot instance segmentation (FSIS). While we argue the FSL paradigm should be effective as well for addressing instance segmentation with limited data, it is by no means trivial to couple the two practically. Crucial to any FSL approach is an appropriate mechanism for encoding and utilizing the support set to guide the base network (\textit{e.g.}, ResNet~\cite{he2016deep} for classification or FCN~\cite{long2015fully} for semantic segmentation). In comparison with the tasks of visual classification or semantic segmentation, designing such a guidance mechanism for instance segmentation becomes far more challenging, which is mainly because instance segmentation networks usually have more complex structures.

In previous attempts~\cite{michaelis2018one, yan2019metarcnn}, the authors proposed to establish guided networks upon Mask R-CNN~\cite{he2017mask}, probably the most representative model for general instance segmentation. Mask R-CNN is a two-stage network, where the first-stage region proposal network (RPN) generates class-agnostic object proposals, and the second-stage subnet consists of three heads for classification, bounding-box (bbox) regression and mask segmentation respectively. Previous works achieve guidance by simply introducing a single guidance module at a certain location in Mask R-CNN. Michaelis \textit{et al.} \cite{michaelis2018one} proposed to make Siamese the backbone network in the first stage to encode the guidance from support set. Consequently, all subsequent components for different tasks (including RPN and the three heads) undesirably have to share the same guidance. In \cite{yan2019metarcnn}, guidance is injected into Mask R-CNN at the front of the second stage by taking class-attentive vectors extracted from support set to reweight the feature maps, which enforces all second-stage components to share the same guidance and totally ignores the first-stage RPN.

\begin{figure*}[!htb]
	\centering
	\includegraphics[width = 15.5cm]{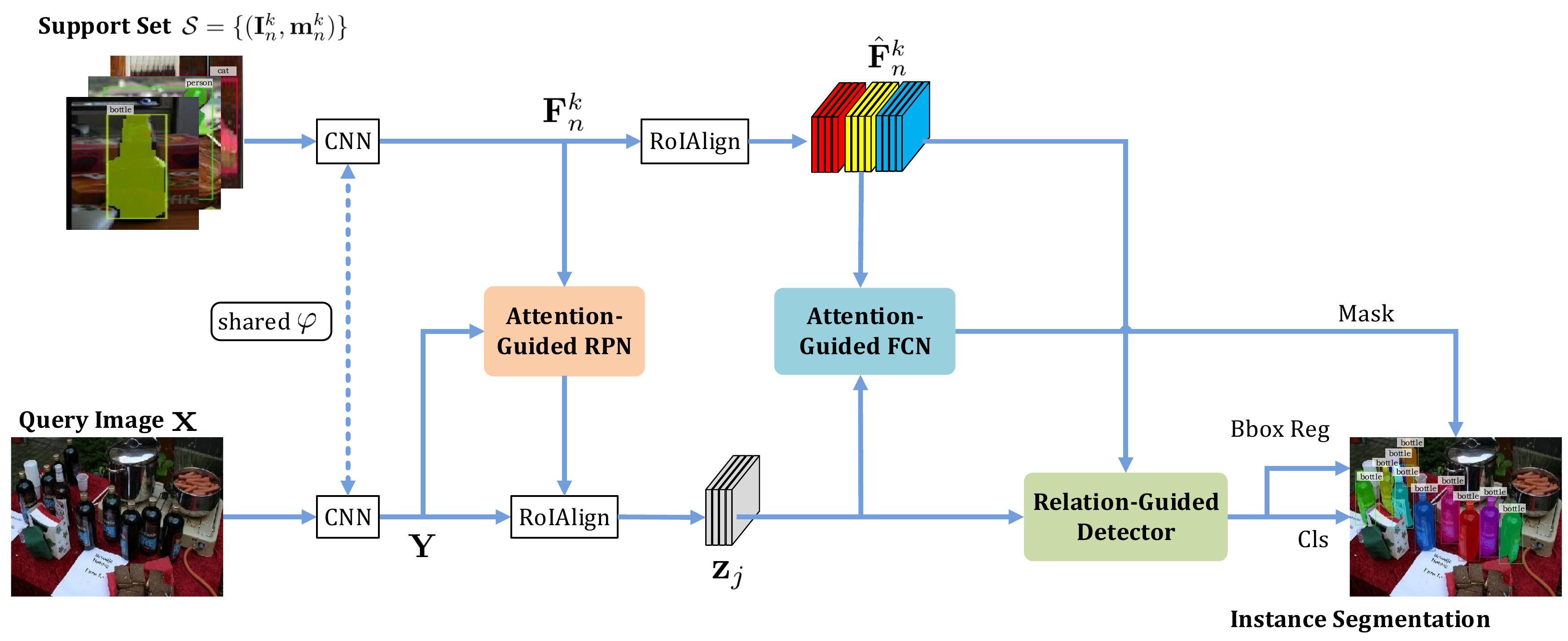}
	\caption{An overview of the proposed Fully Guided Network (FGN). FGN is established upon Mask R-CNN~\cite{he2017mask}, where a support set is encoded and utilized to guide the three key components in Mask R-CNN, through the Attention-Guided RPN (AG-RPN), the Relation-Guided Detector (RG-DET) and the Attention-Guided FCN (AG-FCN), respectively.   }
	\vspace{-1\baselineskip}
\label{fig:fgn-overview}
\end{figure*}

In this paper, we present a Fully Guided Network (FGN) to address few-shot instance segmentation, as conceptually demonstrated in Fig.~\ref{fig:fsis-concept}. FGN conjoins the few-shot learning paradigm with Mask R-CNN to establish a guided network. Different from prior works \cite{michaelis2018one, yan2019metarcnn}, the key philosophy of FGN is that, \textit{components for different tasks in Mask R-CNN should be guided differently to achieve full guidance} (which gives reason to the name of ``Fully Guided Network''). Our intuition is that, the problem setting of FSIS brings different challenges to the various components in Mask R-CNN, which are difficult to be addressed by the use of a single guidance mechanism. Towards this end, FGN introduces three guidance mechanisms into Mask R-CNN, namely, the Attention-Guided RPN (AG-RPN), the Relation-Guided Detector (RG-DET) and the Attention-Guided FCN (AG-FCN), respectively. AG-RPN encodes the support set by class-aware attention, which is then utilized to guide RPN so that it can focus on the novel classes of concern and generate class-aware proposals. RG-DET guides the detector branch by an explicit comparison scheme to adapt to the inter-class generalization in FSIS. AG-FCN also takes attentional information from the support set to guide the mask segmentation procedure. Specific guidance modules are carefully designed and effective training strategy is suggested for model learning (see Figure~\ref{fig:fgn-overview} and Section~\ref{sec:approach} for details). Experimental results on public datasets demonstrate the proposed FGN can outperform the state-of-the-art FSIS approaches. In summary, the main contributions of our work are two-fold:
\begin{itemize}
\setlength{\itemsep}{0pt}
\item We propose the Fully Guided Network, a novel framework for few-shot instance segmentation.

\setlength{\itemsep}{0pt}
\item We suggest three effective guidance mechanisms, \textit{i.e.,} AG-RPN, RG-DET and AG-FCN, leading to superior performance.
\end{itemize}


\section{Related Work}
In this section, we briefly review the related literature. 

\vspace{2pt}
\textbf{Instance Segmentation.} Instance segmentation can be viewed as a task at the intersection of semantic segmentation and object detection, which has made significant advances in recent years~\cite{hariharan2014simultaneous, he2017mask,ren2017end,hayder2017boundary,liu2018path, chen2018masklab, chen2019hybrid}, benefited from deep CNN. Existing instance segmentation approaches are either proposal-based or proposal-free. The most representative work of the former category may be Mask R-CNN~\cite{he2017mask}, which utilizes an RPN to generate class-independent object candidates in the first stage, and the second-stage procedure deals with these candidates only. Other influential works include~\cite{huang2019mask, liu2018path, chen2019hybrid}. The latter category of methods directly performs instance segmentation without relying on RPN, to balance between performance and computational efficiency. Representative works include~\cite{liang2017proposal, fu2019retinamask}. Instance segmentation has been mainly explored under the fully supervised setting so far, which may be impractical for certain applications.

\textbf{Few-Shot Classification.} FSL~\cite{vinyals2016matching,snell2017prototypical} has recently emerged as a promising paradigm for learning predictive models from very limited training data (typically a handful of training samples only for each class). An external dataset with a large number of labeled data (but of different classes from the target ones) is usually necessitated, from which a set of episodes are sampled to simulate the target task. A conditional classifier is then learned from these episodes, which makes predictions conditioned on a support set. The conditional classifier is expected to be generalized well to the target task (on novel classes). A number of few-shot classification models have been proposed recently, including Matching Networks~\cite{vinyals2016matching}, Prototypical Networks~\cite{snell2017prototypical}, Relation Networks~\cite{sung2018learning}, the models based on Siamese CNN~\cite{koch2015siamese}, graph CNN~\cite{garcia2018few}, \textit{etc}. These models can be distinguished by how they encode and utilize the support set to guide the base network.  

\textbf{Few-Shot Semantic Segmentation.} It is natural to consider adapting 
the FSL paradigm to other computer vision tasks, like semantic segmentation, object detection, \textit{etc}. In light of the spirit of few-shot classification, Shaban~\textit{et al.} ~\cite{boots2017one} proposed to utilize a conditioning branch to encode the support set and modulate an FCN-based segmentation branch to achieve one-shot semantic segmentation. Following a similar structure, some authors suggested different schemes for encoding the support set or for imposing modulation on the segmentation branch~\cite{rakelly2018conditional, zhang2018sg, dong2018few}.

\textbf{Few-Shot Object Detection.} It is more challenging to adapt FSL to object detection (termed as few-shot object detection) since object detection requires localization. Some works address this problem from the perspectives of self-paced learning~\cite{dong2018few} or transfer learning~\cite{chen2018lstd}. In~\cite{schwartz2018repmet}, Schwartz \textit{et al.} proposed to integrate a representative-based metric learning approach with the Faster R-CNN framework. In~\cite{kang2019few}, Kang \textit{et al.} presented a conditioned YOLO framework~\cite{redmon2016you} with reweighted features for few shot object detection. These methods can only yield object bounding boxes, rather than instance masks.

Most closely related to ours, the works in~\cite{michaelis2018one, yan2019metarcnn} consider FSIS by constructing guided networks upon Mask R-CNN. However, the overall performance is still limited, possibly due to the fact that, guidance driven by the support set cannot fully affect the base network as aforementioned.  More effective guidance mechanisms for FSIS  largely remain  to be explored.

\section{Approach}
\label{sec:approach}
In this section, we start with the problem statement of few-shot instance segmentation. Then we describe the proposed Fully Guided Network, followed by the strategy for model training. 
\subsection{Problem Statement}
\label{subsec:ps}
Suppose for a set of \textit{base classes} $\mathcal{C}^{\text{base}}$, we have a large set of images annotated with object instances, denoted by $\mathcal{D}^{\text{base}}$. Now let us consider a different set of semantic classes $\mathcal{C}^\text{novel}$ (called \textit{novel classes}),  which do not overlap with the base classes, \textit{i.e.}, $\mathcal{C}^{\text{base}} \cap \mathcal{C}^{\text{novel}} = \phi$. For these novel classes, we only have a very limited number of annotated instances $\mathcal{D}^\text{novel}$, referred to as \textit{support set}. In practice, this is usually due to difficulties in collecting images or acquiring instance-level annotations. The task of \textit{few-shot instance segmentation (FSIS)} is to segment, from any given \textit{query image} $\mathbf{I}^q$, all the object instances belonging to the novel classes. Note that when $|\mathcal{C}^\text{novel}| = N$ ($|\cdot|$ represents the cardinality of a set throughout this paper) and there are $K$ annotated instances for each novel class, we call it an \textit{$N$-way $K$-shot instance segmentation} task.

In this paper, we conjoin the few-shot learning paradigm with general instance segmentation to address the FSIS problem. Following the spirit of few-shot classification~\cite{vinyals2016matching,snell2017prototypical}, we simulate a quantity of $N$-way $K$-shot instance segmentation tasks $\mathcal{T} = \{ (\mathcal{S}_i, \mathbf{x}_i)\}_{i=1}^{|\mathcal{T}|}$ by randomly sampling support sets and queries from $\mathcal{D}^{\text{base}}$ (of the base classes $\mathcal{C}^{\text{base}}$), where the $i$-th task is formed by sampling a support set $\mathcal{S}_i$ and a query image $\mathbf{x}_i$. By the use of these simulated tasks $\mathcal{T}$, we learn a conditional instance segmentation model $f_{\theta}(\mathbf{x}|\mathcal{S})$ parameterized by $\theta$, which performs instance segmentation on the query image $\mathbf{x}$ conditioned on the support set $\mathcal{S}$. The learned model $f_{\theta}(\mathbf{x}|\mathcal{S})$ can then be applied to the target task, \textit{i.e.}, $N$-way $K$-shot instance segmentation over the novel classes $\mathcal{C}^\text{novel}$ (simply letting $\mathcal{S} = \mathcal{D}^\text{novel}$ and $\mathbf{x} = \mathbf{I}^q$). It is worth pointing out that, instead of straightforwardly learning $f_{\theta}(\mathbf{x})$, our strategy is to learn a conditional model $f_{\theta}(\mathbf{x}|\mathcal{S})$, which can be viewed as to utilize the support set $\mathcal{S}$ to guide the instance segmentation of $\mathbf{x}$. The presence of guidance plays a critical role for the model trained on the base classes $\mathcal{C}^{\text{base}}$ to generalize well to the novel classes $\mathcal{C}^{\text{novel}}$. 

\begin{figure}[tb]
	\centering
	\includegraphics[width = 7.5cm]{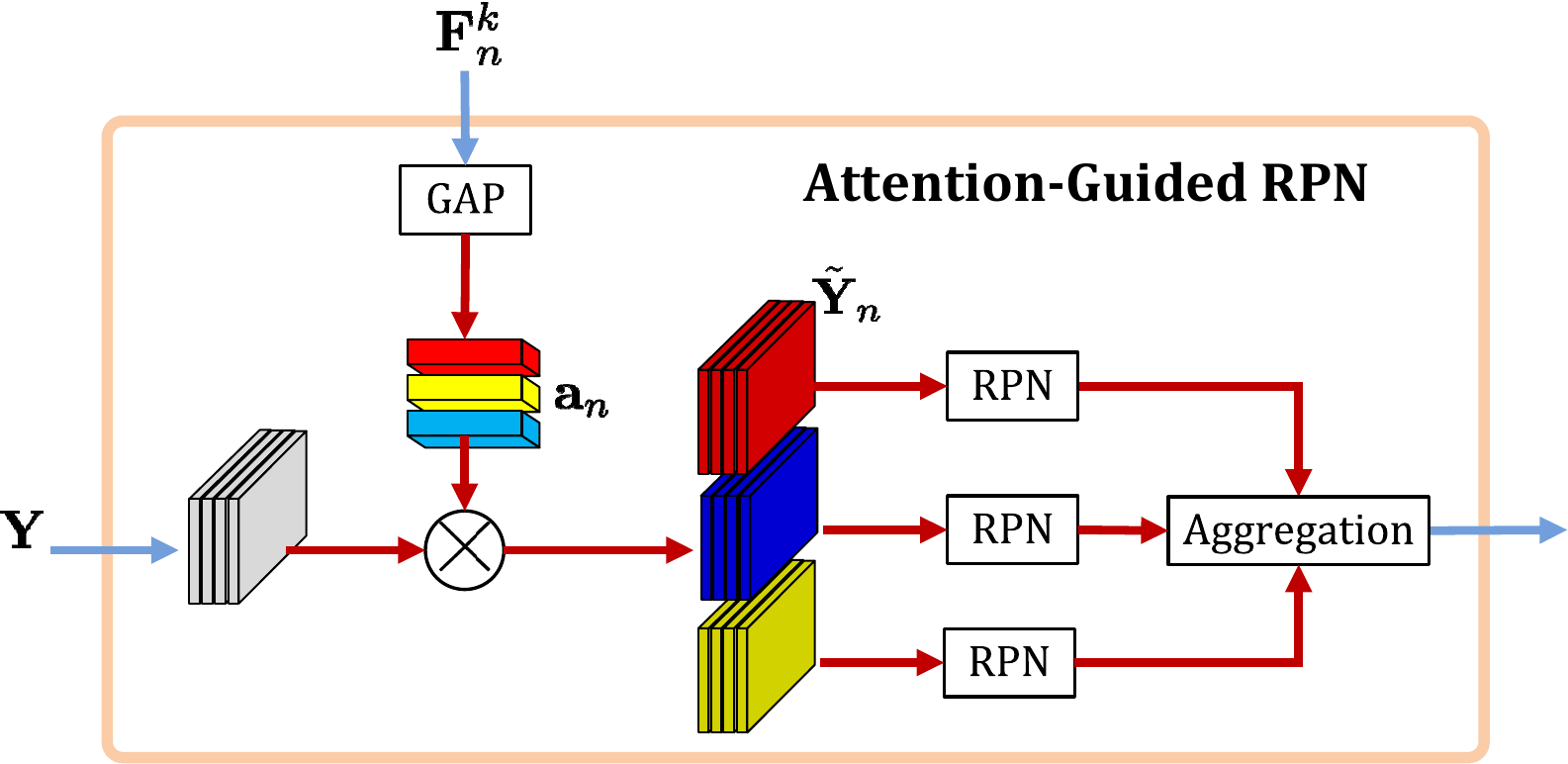}
	\caption{The structure of Attention-Guided RPN (AG-RPN).}
	\vspace{-1\baselineskip}
\label{fig:ag-rpn}
\end{figure}

\subsection{Fully Guided Network}
Central to any FSIS approach is how to effectively encode and utilize the support set to guide the basic instance segmentation network (mostly typically Mask R-CNN~\cite{he2017mask}). Previous works fulfill such guidance by incorporating a single guidance module at a certain location in Mask R-CNN, which may undesirably enforce components for different tasks to share the same guidance~\cite{vinyals2016matching}, or neglect certain components~\cite{snell2017prototypical}. We present the Fully Guided Network (FGN) in this paper, which is distinct from previous works~\cite{vinyals2016matching,snell2017prototypical} in that, components for different tasks in Mask R-CNN are guided by the support set differently to achieve full guidance.

An overview of the proposed FGN is demonstrated in Fig.~\ref{fig:fgn-overview}. Generally, FGN introduces guidance into Mask R-CNN at three key components, \textit{i.e.}, the RPN, the detection branch (including classification and bbox  regression) and the mask branches, leading to the Attention-Guided RPN (AG-RPN), the Relation-Guided Detector (RG-DET) and the Attention-Guided FCN (AG-FCN), respectively. In the proposed FGN, the given support set $\mathcal{S}$ (containing $K$ annotated instances for each of the $N$ classes) and the query image $\mathbf{x}$ are encoded by a shared backbone $\varphi$ (ResNet101~\cite{he2016deep} in our implementation) to give the feature maps $\mathbf{F}_n^k, \ \mathbf{Y}$  $\in \mathbb{R}^{H \times W \times C}$ respectively. $\mathbf{F}_n^k$ encodes the support set, which is used by AG-RPN to guide the proposal generation from $\mathbf{Y}$ in the first stage. Then, in the second stage, for each proposal [also called {Region-of-Interest (RoI)}] with the aligned feature maps $\mathbf{z}_j \in \mathbb{R}^{h \times w \times C}$, the aligned $\hat{\mathbf{F}}_n^k \in \mathbb{R}^{h \times w \times C}$ is utilized by RG-DET to guide the classification and bbox heads, and by AG-FCN to guide the mask head. Another key contribution of our work is to design novel and effective guidance mechanisms for these modules, which are detailed as below.

\textbf{Attention-Guided RPN.} Mask R-CNN relies on RPN to obtain class-agnostic proposals of potential objects for subsequent processing. Under the problem setting of FSIS, RPN has to be trained on the base classes $\mathcal{C}^\text{base}$ and tested on a solely different set of novel classes $\mathcal{C}^\text{novel}$. In this case, RPN may generate a lot of undesired proposals but miss the ones of concern, especially when $\mathcal{C}^\text{novel}$ departs far from $\mathcal{C}^\text{base}$, or the number of novel classes is small, which will largely degrade overall performance. To tackle this issue, our idea is to introduce guidance from the support set into RPN such that it can focus on the classes of concern and generate class-aware proposals, which we call \textit{Attention-Guided RPN (AG-RPN)}.

The structure of AG-RPN is depicted in Fig.~\ref{fig:ag-rpn}.  The feature maps $\mathbf{F}_n^k \in \mathbb{R}^{H \times W \times C}$ with $ n = 1, ..., N, k = 1, ..., K$, which encode the support set, undergo the global average pooling (GAP) and the averaging operation over each individual class, given by 

\begin{equation}
\mathbf{a}_{n} = \frac{1}{K} \sum_{k=1}^K \text{GAP} \left( \mathbf{F}_n^k \right),  \quad n = 1, ..., N, 
\label{eq:cav}
\end{equation}
with $\{\mathbf{a}_1, ..., \mathbf{a}_N\} \in \mathbb{R}^{C \times 1} $ being the \textit{class-attentive vectors} associated with the $N$ novel classes. Each $\mathbf{a}_n$ is then taken to weight the feature maps of the query image $\mathbf{Y} \in \mathbb{R}^{H \times W \times C}$ as below

\begin{equation}
\tilde{\mathbf{Y}}_{n} = \mathbf{Y} \otimes \mathbf{a}_n,  \quad n = 1, ..., N,  
\label{eq:otimes}
\end{equation}
which means taking $\mathbf{a}_n$ to perform element-wise multiplication along the channel dimension at every spatial location in $\mathbf{Y}$. Each $\tilde{\mathbf{Y}}_{n}$ is fed into the basic RPN for proposal generation independently and the results are then aggregated to yield the final proposals. The aggregation procedure can be described as follows: For each particular anchor, an objectness score can be acquired through the RPN over every $\tilde{\mathbf{Y}}_{n}$, and the softmax results over the $N$ scores are taken as the class-aware confidence of the anchor. {Anchor refinement is conducted by the regression corresponding to the top matching score during inference.} The final proposals are picked up from the anchors by thresholding their confidence and performing non-maximal suppression.

\begin{figure}[!t]
	\centering
	\includegraphics[width = 8.3cm]{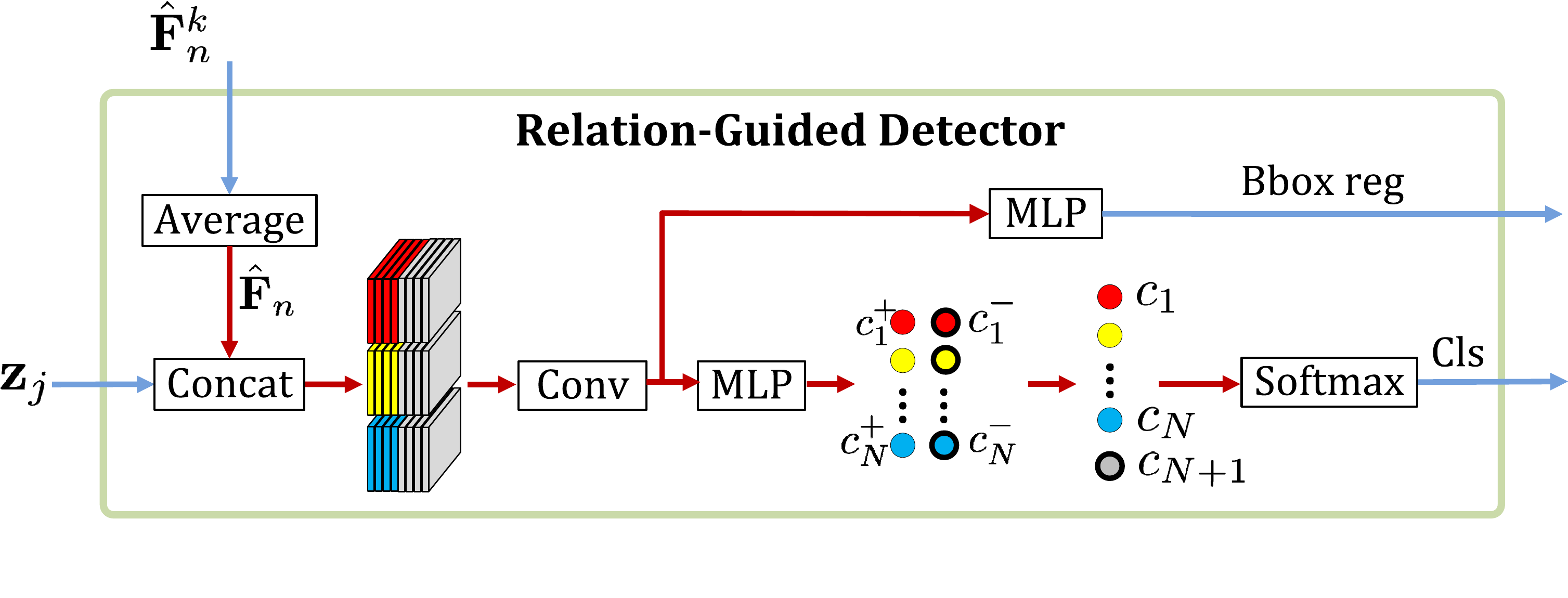}
	\caption{The structure of Relation-Guided Detector (RG-DET).}
\label{fig:rg-det}
\end{figure}

\vspace{4pt}
\textbf{Relation-Guided Detector.} The guidance on the detector branch in Mask R-CNN (including the classification and bbox regression heads) is imposed in an implicit way in previous works~\cite{michaelis2018one, yan2019metarcnn}, which just simply modulate the feature extraction in the first or second stage by the use of support set. In this paper, we propose a different guidance mechanism for the detector (actually the classification branch), termed as \textit{Relation-Guided Detector (RG-DET)}. RG-DET achieves guidance by explicitly comparing the features extracted from the support set and the RoI, inspired by the Relation Network (RN)~\cite{sung2018learning} originally proposed for few-shot classification. We favor RN mainly because it is characterized by that, both the feature embedding and the similarity measure are learnable, compared to other competitors like~\cite{vinyals2016matching,snell2017prototypical,koch2015siamese}. 

Unfortunately, RN cannot be directly deployed to our task because there exists an essential difference between our problem here and the general few-shot classification, that is, the rejection of background class. RG-DET operates on individual RoIs output by AG-RPN, which may inevitably contain background RoIs belonging to neither of the novel classes in the support set. By contrast, recall that few-shot classification methods (including RN) always classifies the query to be one of the classes indicated by the support set. Taking into account the background rejection issue, the structure of RG-DET is illustrated in Fig.~\ref{fig:rg-det}.

For a particular RoI, its aligned feature maps $\mathbf{z}_j \in \mathbb{R}^{h \times w \times C} $ are concatenated with the $N$ aligned feature maps $\hat{\mathbf{F}}_n = \left( \frac{1}{K} \sum_{k}\hat{\mathbf{F}}_n^k \right) \in \mathbb{R}^{h \times w \times C}$ extracted from the support set (as shown in Fig.~\ref{fig:rg-det}), followed by a stack of conv and fc layers (termed as MLP), to give the matching scores (the cls branch) and the object box (the bbox reg branch). The matching score between  $\mathbf{z}_j$ and the $i$-th feature maps $\hat{\mathbf{F}}_n$ is represented by a doublet $(c_i^+, c_i^-)$, where $c_i^+$ and $c_i^-$ stand for the confidence of matching the $i$-th class and the background respectively. To enable background rejection, we need to derive an $(N\!+\!1)$-length matching vector $\mathbf{c} = (c_1, ..., c_N, c_{N+1})$ from the $2N$ original scores, with $c_i, i = 1, ..., N$ reflecting the confidence of the $i$-th class and $c_{N+1}$ the background. For this purpose, we set $c_i = c_i^+$ and $c_{N+1} = c_{i^*}^-$ with $i^* = \argmax_i \left\{c_i^+\right\}$, which physically means we depend on the best-matched class (the most reliable one) to estimate the confidence of background $c_{N+1}$. A softmax operation is then performed over the matching vector $\mathbf{c}$, yielding the final classification score. 

The bbox regression branch shares the concatenation and the first conv layer with the classification branch, but has a separate MLP layer as shown in Fig.~\ref{fig:rg-det}. 

\begin{figure}[!t]
	\centering
	\includegraphics[width = 5.5cm]{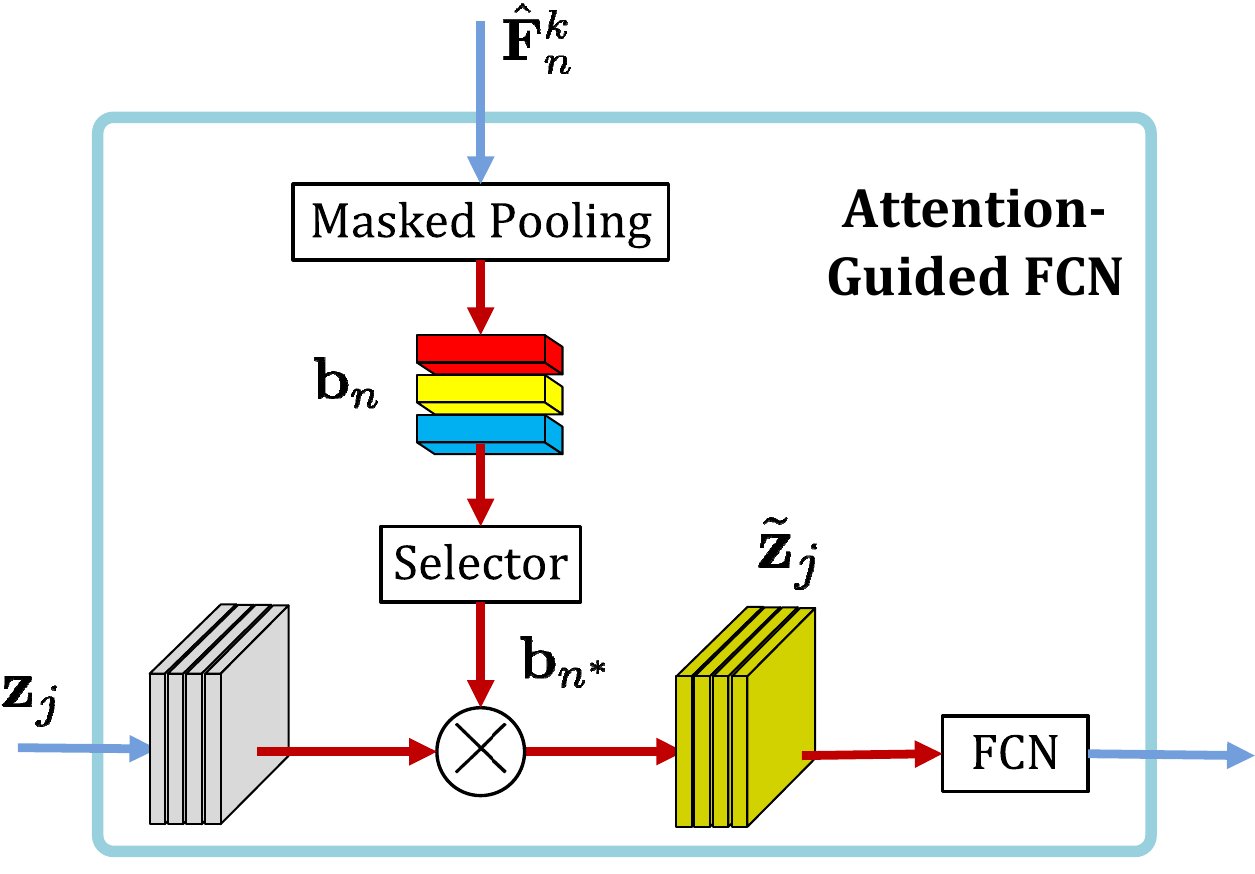}
	\caption{The structure of Attention-Guided FCN.}
	\vspace{-1\baselineskip}
\label{fig:ag-fcn}
\end{figure}

\vspace{4pt}
\textbf{Attention-Guided FCN.}  As illustrated in Fig.~\ref{fig:ag-fcn}, the \textit{Attention-Guided FCN (AG-FCN)} introduces guidance into the FCN-based mask head. AG-FCN basically follows the guidance scheme for few-shot semantic segmentation~\cite{shaban2017one}, except for two modifications. First, an operation of \textit{masked pooling}~\cite{zhang2018sg} is performed on the aligned feature vectors $\hat{\mathbf{F}}_n^k \in \mathbb{R}^{h \times w \times C}$ before computing the class-attentive vectors $\{\mathbf{b}_1, ..., \mathbf{b}_N\} \in \mathbb{R}^{C \times 1}$ as described in Eq.~(\ref{eq:cav}). Masked pooling on $\hat{\mathbf{F}}_n^k$ means pooling $\hat{\mathbf{F}}_n^k$ within the binary mask $\hat{\mathbf{m}}_n^k \in \mathbb{R}^{h \times w \times C}$, which is obtained by performing RoIAlign over the original instance mask ${\mathbf{m}}_n^k \in \mathbb{R}^{H \times W \times C}$. Second, a \textit{selector} is used to pick up the one $\mathbf{b}_{n^*}$ from $\{\mathbf{b}_1, ..., \mathbf{b}_N\}$, where $n^*$ is chosen to be the ground truth class for training, and the one with the highest classfication score for testing. Note that $\tilde{\mathbf{z}}_j = \mathbf{z}_j \otimes \mathbf{b}_{n^*}$ where the operator $\otimes$ is identical to that in Eq.~(\ref{eq:otimes}). 

\subsection{Training Strategy}
\label{subsec:trainstr}
FGN is a two-stage structure since it is based on Mask R-CNN. Hence, our pipeline for training is basically similar to Mask R-CNN (including the loss functions). But differently, following the common practice in~\cite{chen2018lstd,kang2019few,yan2019metarcnn}, our training includes two steps. For the first step, we purely take $\mathcal{D}^{\text{base}}$ of the base classes $\mathcal{C}^{\text{base}}$ as the training data. And for the second step, we take data from both the base classes and the novel classes, \textit{i.e.}, $\mathcal{C}^{\text{base}} \cup \mathcal{C}^{\text{novel}}$, to further fine-tune the model. More precisely, the second-step training data consist of the whole support set $\mathcal{D}^{\text{novel}}$ (containing $NK$ instances) and $3K$ instances for each class in $\mathcal{C}^{\text{base}}$ randomly sampled from $\mathcal{D}^{\text{base}}$, which contain totally $(N+3|\mathcal{C}^{\text{base}}|)K$ instances. Our training requires randomly sampling the training set to simulate the target FSIS tasks (constructing the episodes), which will be detailed in Section~\ref{subsec:expset}.

\section{Experiments and Results}
In this section, we present experimental results to evaluate the effectiveness of our method, mainly including: 1) comparison with the state-of-the-art methods; 2) ablation study with several variant baselines. Our method was implemented in TensorFlow and Keras on a workstation with 4 NVIDIA Titan XP GPUs.  

\begin{table*}[]
\small
    \centering
    \begin{tabular}{c !{\vrule width1.2pt} c|c|c !{\vrule width1.2pt} c|c|c}
      \Xhline{1pt}
      \multirow{2}{*}{\textbf{Methods}} &
      \multicolumn{3}{c !{\vrule width1.2pt} }{\textbf{Segmentation}}  &
      \multicolumn{3}{c}{\textbf{Detection}}\\\cline{2-7}
      & \textbf{1way-1shot} &  \textbf{3way-1shot} & \textbf{3way-3shot} & \textbf{1way-1shot} &  \textbf{3way-1shot} & \textbf{3way-3shot} \\\Xhline{1pt}
      MRCNN-FT & 0.4 & 0.5 & 2.7 & 6.0 & 5.2 & 10.2 \\\hline
      Siamese MRCNN~\cite{michaelis2018one} & 13.8 & 6.3 &  6.6 & 23.9 & 11.5 & 13.3 \\\hline
      Meta R-CNN~\cite{yan2019metarcnn} & 12.5 & 12.1 & 15.3 & 20.1 & 19.2 & 23.4  \\\hline
      FGN & \textbf{16.2} & \textbf{13.0} & \textbf{17.9} & \textbf{30.8} & \textbf{23.5} & \textbf{32.9} \\\Xhline{1pt}
    \end{tabular}
    \caption{Performance in terms of $\text{mAP}_\text{50}$ obtained by various methods under the \textbf{COCO2VOC} setting. Both the segmentation and detection results are reported for comparison. }
    \label{tab:perf_novel_voc}
\end{table*}

\begin{table*}[]
\small
    \centering
    \begin{tabular}{c !{\vrule width1.2pt} c|c|c !{\vrule width1.2pt} c|c|c}
      \Xhline{1pt}
      \multirow{2}{*}{\textbf{Methods}} &
      \multicolumn{3}{c !{\vrule width1.2pt} }{\textbf{Segmentation}}  &
      \multicolumn{3}{c}{\textbf{Detection}}\\\cline{2-7}
      & \textbf{1way-1shot} &  \textbf{3way-1shot} & \textbf{3way-3shot} & \textbf{1way-1shot} &  \textbf{3way-1shot} & \textbf{3way-3shot} \\\Xhline{1pt}
      MRCNN-FT & 25.3 & 25.0 & 27.4 & 27.3 & 27.1 & 29.7 \\\hline
      Siamese MRCNN~\cite{michaelis2018one} & 24.2 & 8.8 & 9.1 & 26.4 & 9.7 & 10.1 \\\hline
      Meta R-CNN~\cite{yan2019metarcnn} & 14.9 & 14.1 & 15.2 & 18.5 & 17.8 & 19.3  \\\hline
      FGN & 24.2 & 13.2 & 14.3 & 27.2 & 16.7 & 17.3 \\\Xhline{1pt}
    \end{tabular}
    \caption{Addition experimental results to demonstrate the challenges of the FSIS problem setting. In this experiment, the settings of $\mathcal{C}^{\text{base}}$ and $\mathcal{D}^{\text{base}}$ are identical to those in COCO2VOC, but $\mathcal{C}^{\text{novel}} \subset \mathcal{C}^{\text{base}}$ and the testing tasks are sampled from COCO's validation set. }
    \label{tab:perf_base_coco}
\end{table*}

\subsection{Experimental Settings}
\label{subsec:expset}

We adopt two commonly-used datasets for our experiments, \textit{i.e.}, Microsoft COCO 2017~\cite{lin2014microsoft} and PASCAL VOC 2012~\cite{everingham2010pascal} (termed as \textbf{COCO} and \textbf{VOC} respectively). {COCO} has $80$ object classes, consisting of a training set ({trainset}) with $118,287$ images and a validation set ({valset}) with $4,952$ images. {VOC} covers $20$ classes that are a subset of COCO's $80$ classes, with a trainset of $1,464$ images (annotated with instance masks) and a valset of $1,449$ images. 

\vspace{2pt}
\textbf{General Settings.} According to the problem definition in Section~\ref{subsec:ps}, our evaluation requires the following basic settings: \textit{ 1) Setting the base classes $\mathcal{C}^\text{base}$ and the novel classes $\mathcal{C}^\text{novel}$, and accordingly the training set $\mathcal{D}^\text{base}$ and the query set $\mathcal{D}^\text{novel}$ (testing set):} As our main setting, we adopt a challenging cross-dataset setting to better compare the generalization ability of various models, inspired by previous works~\cite{kang2019few, yan2019metarcnn}. Specifically, we set the $20$ classes at the intersection of {COCO} and {VOC} to be $\mathcal{C}^\text{novel}$ and the rest $60$ classes covered by {COCO} but not {VOC} to be $\mathcal{C}^\text{base}$. Further, we take from {COCO}'s trainset the subset belonging to $\mathcal{C}^\text{base}$ as the training set $\mathcal{D}^\text{base}$, and take {VOC}'s valset (belonging to the $20$ novel classes $\mathcal{C}^\text{novel}$) to construct the testing set (see details later). We refer to this main experimental setting as \textbf{COCO2VOC}. Additionally, we also consider another setting termed as \textbf{VOC2VOC}, which only uses the {VOC} dataset. More precisely, we randomly sample $15$ out of $20$ classes covered by {VOC} to be the base classes $\mathcal{C}^\text{base}$ and the rest $5$ are taken as $\mathcal{C}^\text{novel}$. The training set $\mathcal{D}^\text{base}$ and the query set $\mathcal{D}^\text{novel}$ are constructed respectively from {VOC}'s trainset and valset. \textit{ 2) Specifying the numbers of $N$ and $K$:} We consider three different settings (a) $N = 1, K = 1$ (termed as \textbf{1way-1shot}); (b) $N = 3, K = 1$ (termed as \textbf{3way-1shot}); (c) $N = 3, K = 3$ (termed as \textbf{3way-3shot}).
 
\textbf{Methods for Comparison.} To our knowledge, there exist only two FSIS methods in the literature so far, \textit{i.e.}, \textbf{Siamese MRCNN}~\cite{michaelis2018one} and \textbf{Meta R-CNN}~\cite{yan2019metarcnn}, which are included in our comparison. Similar to our FGN, {Siamese MRCNN}
and {Meta R-CNN} also achieve FSIS by introducing guidance into Mask R-CNN (but using different guidance mechanisms), for which we use the source codes released by the authors for our experiments. Besides, we also build a baseline for comparison, termed as \textbf{MRCNN-FT}, which is basically a Mask R-CNN trained with the strategy detailed in Section~\ref{subsec:trainstr}. 

\textbf{Implementation Details.} We  follow the training strategy in Section~\ref{subsec:trainstr} and the settings of  $\{\mathcal{C}^\text{base}, \mathcal{D}^\text{base}, \mathcal{C}^\text{novel},\mathcal{D}^\text{novel}, N, K\}$ above in Section~\ref{subsec:expset} to train our FGN model. We use ResNet101~\cite{he2016deep} as the backbone for our model. The initial learning rates of SGD for training the first-stage AG-RPN and the second-stage RG-DET and AG-FCN are set to $0.01$ and $0.001$ respectively. We train for $60, 000$ steps and a $10$-times learning rate decay is applied to the second-half steps.

To construct the simulated tasks $\mathcal{T} = \{ (\mathcal{S}_i, \mathbf{x}_i)\}_{i=1}^{|\mathcal{T}|}$ (typically called ``episodes'') for training, we basically follow the sampling strategy proposed in~\cite{vinyals2016matching}. Note that, we crop the local patches extended by $20$ pixels around ground truth boxes of instances to form the support set, rather than using holistic images. And for testing, the tasks $\{ (\mathcal{D}^\text{novel}_i, \mathbf{I}^q_i)\}_{i} $ are constructed to ensure every novel class in every image in the testing set is tested for once. Specifically, for each image $\mathbf{I}^q_i$, we collect all the classes it covers. Then, for each class we randomly sample other $N-1$ classes and pick up instances accordingly to form an $N$-way $K$-shot episode. We report the average performance over all the testing tasks.


\begin{figure*}[!htb]
	\centering
	\includegraphics[width = 15.5cm]{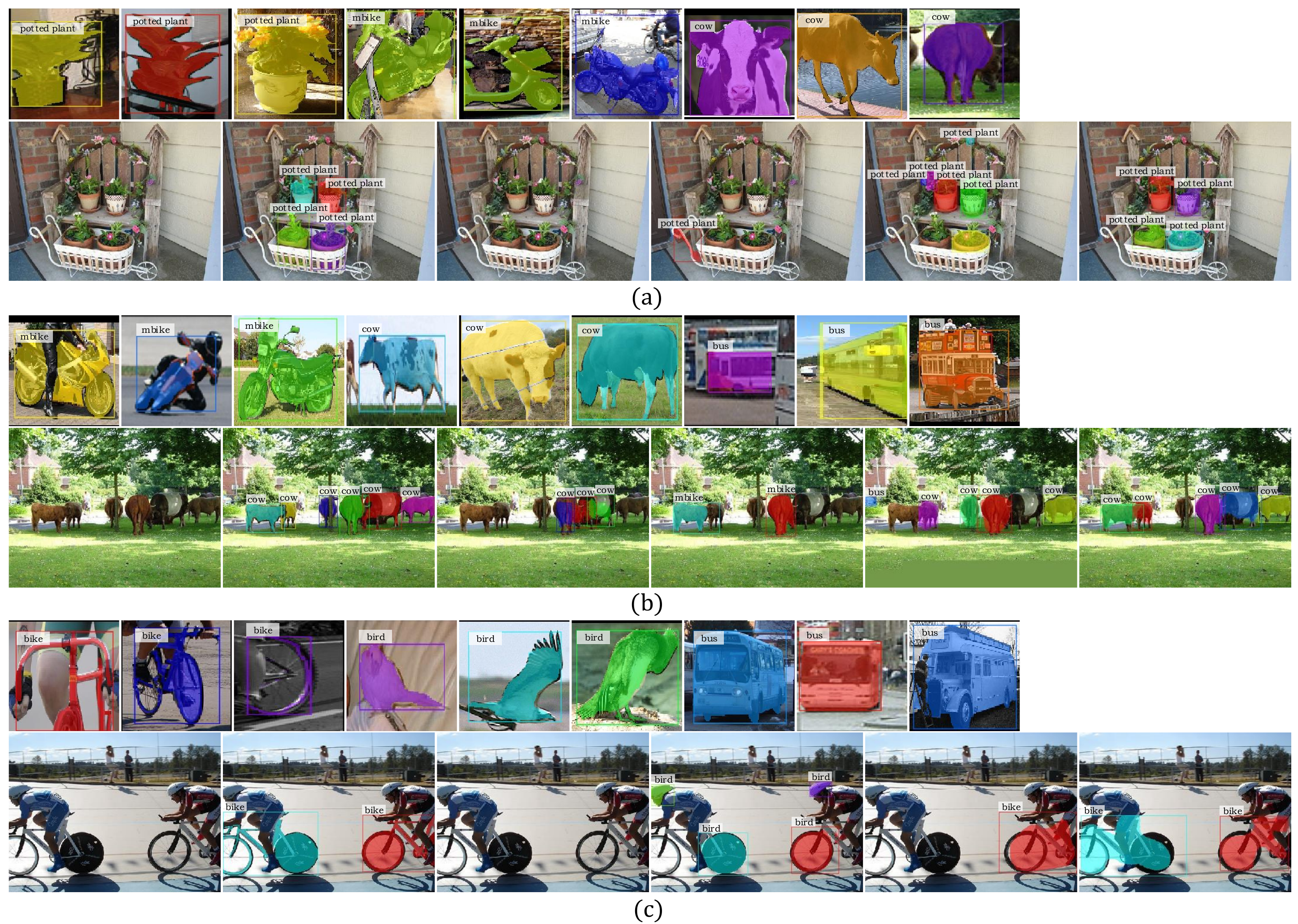}
	\caption{Exemplary results obtained by various results under the \textbf{COCO2VOC 3way-3shot} setting. In each group (a) - (c), the images in the top row are the support set. And in the bottom row, from left to right are the query image, the ground truth, and the results obtained by MRCNN-FT, Siamese MRCNN~\cite{michaelis2018one}, Meta R-CNN~\cite{yan2019metarcnn} and our FGN.}
\label{fig:qual_results}
\end{figure*}

\begin{table*}[t!]
\small
    \centering
    \begin{tabular}{c !{\vrule width1.2pt} c|c|c !{\vrule width1.2pt} c|c|c}
      \Xhline{1pt}
      \multirow{2}{*}{\textbf{Methods}} &
      \multicolumn{3}{c !{\vrule width1.2pt} }{\textbf{Segmentation}}  &
      \multicolumn{3}{c}{\textbf{Detection}}\\\cline{2-7}
      & \textbf{1way-1shot} &  \textbf{3way-1shot} & \textbf{3way-3shot} & \textbf{1way-1shot} &  \textbf{3way-1shot} & \textbf{3way-3shot} \\\Xhline{1pt}
      Siamese MRCNN~\cite{michaelis2018one} & 8.2 & 4.4 & 5.2 & \textbf{17.9} & 8.7 & 9.0 \\\hline
      Meta R-CNN~\cite{yan2019metarcnn} & 4.2 & 3.6 & 7.3 & 8.0 & 7.3 & 14.4 \\\hline
      FGN & \textbf{8.4} & \textbf{7.3} & \textbf{9.6} & 15.4 & \textbf{11.3} & \textbf{16.2} \\\Xhline{1pt}
    \end{tabular}
    \caption{Performance in terms of $\text{mAP}_\text{50}$ obtained by various methods under the \textbf{VOC2VOC} setting. Both the segmentation and detection results are reported for comparison. }
    \label{tab:perf_voc}
\end{table*}

\subsection{Results}

We present the main results under the settings of {COCO2VOC} and {VOC2VOC} and related analysis respectively in the following. 

\textbf{COCO2VOC.} The FSIS performance obtained by the various methods under the {COCO2VOC} setting is comparatively reported in Table~\ref{tab:perf_novel_voc}, where we use $\text{mAP}_\text{50}$ as the quantitative performance measure. As can be observed that, our FGN  can generally outperform the two state-of-the-art methods {Siamese MRCNN}~\cite{michaelis2018one} and {Meta R-CNN}~\cite{yan2019metarcnn} to a large margin for the three settings of $N$ and $K$. Siamese MRCNN ~\cite{michaelis2018one} performs comparatively to ours in case of 1way-1shot, but degrades heavily under the other two settings. This is probably because that, the guidance in this approach follows the Siamese Network mechanism which is originally designed for pairwise input. Meta R-CNN~\cite{yan2019metarcnn} does not perform well either, probably because this method relies much on the finetuning procedure in training, which cannot acquire sufficient data for finetuning when $N$ and $K$ are small like in our settings. As expected, the baseline MRCNN-FT performs very poorly, which suggests that the strategy of naively finetuning a model pretained from base classes with data from novel classes is inappropriate for FSIS. 

\begin{table*}[t!]
\small
    \centering
    \begin{tabular}{c|c c c|c|c}
        \Xhline{1pt}
        & AG-RPN & RG-DET & AG-FCN & \textbf{Segmentation} & \textbf{Detection} \\
        \Xhline{1pt}
        FGN-P & \checkmark &  &   & 13.7 & 23.8\\\hline
        FGN-DS & &\checkmark  &\checkmark   & 15.1 & 26.8\\\hline
        FGN-PS &\checkmark  & &\checkmark  & 15.6 & 24.8 \\\hline
        FGN-PD &\checkmark  &\checkmark  &  & 15.1 & 29.1\\\hline
        FGN (Ours) &\checkmark  &\checkmark  &\checkmark   & \textbf{17.9} & \textbf{32.9}\\
        \Xhline{1pt}
    \end{tabular}
    \caption{Ablation study on the effectiveness of full guidance. Comparison among the variants of FGN in terms of $\text{mAP}_\text{50}$.}
    \vspace{-\baselineskip}
    \label{tab:abl-full-guidance}
\end{table*}

In addition to segmentation, we also compare the various methods on the task of few-shot object detection, as shown in Table~\ref{tab:perf_novel_voc}. Our FGN can also outperform the other methods consistently for all the settings. One can further observe that, there is an obvious performance drop from detection to segmentation for all the methods, which may indicate that FSIS cannot be achieved by trivial extension of few-shot object detection methods. We also provide some exemplary results obtained by various methods for visual comparison in Fig.~\ref{fig:qual_results}.

\begin{table}[t!]
\small
    \centering
    \begin{tabular}{ c c c}
        \Xhline{1pt}
          \textbf{RPN} & \textbf{AG-RPN-v1} & \textbf{AG-RPN} \\\Xhline{1pt}
        64.5 & 74.8 & \textbf{92.5} \\\Xhline{1pt}
    \end{tabular}
    \caption{Comparison among the variants of AG-RPN in terms of $\text{AR}_\text{50}$.}
    \vspace{-\baselineskip}
    \label{tab:abl-ag-rpn}
\end{table}

While the proposed FGN can outperform the state-of-the-art as stated above, one may be concerned with a fact that, the performance of various methods (including ours) generally looks limited, significantly worse than conventional instance segmentation. We argue this is likely due to the intrinsic challenges of the FSIS problem, especially in case of low numbers of ways and shots like ours. To justify this point, we further carry out another experiment where the settings of $\mathcal{C}^{\text{base}}$ and $\mathcal{D}^{\text{base}}$ are identical to those in COCO2VOC, but the novel classes $\mathcal{C}^{\text{novel}} \subset \mathcal{C}^{\text{base}}$ and the testing tasks are sampled from COCO's validation set (the data used for testing are different). Such case where $\mathcal{C}^\text{novel} \subset \mathcal{C}^\text{base}$ does not coincide with the problem definition of FSIS but general instance segmentation. Also, MRCNN-FT is a Mask R-CNN trained by the common strategy described in Section~\ref{subsec:trainstr}, which is shared by all the compared methods (including ours). As shown in Table~\ref{tab:perf_base_coco}, under the setting of general instance segmentation, even the standard Mask R-CNN trained in the same fashion as commonly required by FSIS approaches can only achieve limited performance. This may reflect that, the FSIS problem setting is inherently challenging, and the training strategy adopted by these FSIS methods (including our FGN) is effective in this sense. It is worth noticing that, it is not meaningful to make comparison among the various methods under this experimental setting.  

\textbf{VOC2VOC.} In addition to our main setting of COCO2VOC, we also evaluate under the VOC2VOC setting. The results obtained by various methods in terms of  $\text{mAP}_\text{50}$ are listed in Table~\ref{tab:perf_voc}. Although VOC2VOC shares the same validation set as COCO2VOC, it has a far smaller training set ($\sim 1.4 \text{K}$ in contrast to $\sim 118 \text{K}$ images). As a result, the performance of VOC2VOC is worse than that of COCO2VOC for all the methods. In this case, our FGN can still achieve the best overall performance among the compared methods for both segmentation and detection.


\vspace{-0.5em}
\subsection{Ablation Study}

We perform ablation study to further reveal the merits of our FGN. All the following experiments are conducted under the \textbf{COCO2VOC 3way-3shot} setting.

\textbf{Full Guidance.} One key reason of FGN's effectiveness is that we carefully design three guidance mechanisms, \textit{i.e.}, AG-RPN (P), RG-DET (D) and AG-FCN (S) to achieve full guidance. To verify the contributions of these modules, we construct several variants by disabling one or more modules from the full FGN model. 

The results obtained by these variants in terms of $\text{mAP}_\text{50}$ for segmentation and detection are comparatively reported in Table~\ref{tab:abl-full-guidance}. It can be seen from the degraded performance of these variants that, each module contributes to some extent on both tasks. 

\textbf{AG-RPN.} We compare our AG-RPN with the basic RPN in Mask R-CNN and a variant termed as \textbf{AG-RPN-v1} by evaluating separately the quality of the proposals generated. 
{ AG-RPN-v1 follows the design in ~\cite{michaelis2018one} to achieve guidance.}
As can be observed from Table~\ref{tab:abl-ag-rpn} that, AG-RPN (ours)
obtains the best performance in terms of $\text{AR}_{\text{50}}$.

\begin{table}[t!]
\small
    \centering
    \begin{tabular}{ c c c c}
        \Xhline{1pt}
          \textbf{FCN} & \textbf{AG-FCN-v1} & \textbf{AG-FCN-v2} & \textbf{AG-FCN} \\\Xhline{1pt}
        15.1 & 14.5 & 15.6 & \textbf{17.9} \\\Xhline{1pt}
    \end{tabular}
    \caption{Comparison among the variants of AG-FCN in terms of $\text{mAP}_\text{50}$.}
    \vspace{-\baselineskip}
    \label{tab:abl-ag-fcn}
\end{table}

\textbf{AG-FCN.} We construct two variants of AG-FCN (ours) for comparison, termed as \textbf{AG-FCN-v1} and  \textbf{AG-FCN-v2}. AG-FCN-v1 is the FCN guidance mechanism suggested in~\cite{zhang2018sg}  for the task of semantic segmentation. AG-FCN-v2 tiles the channel attention vectors $\mathbf{b}_{n^*}$ to be of the same size as $\mathbf{z}_j$ and then concatenates them together (see Fig.~\ref{fig:ag-fcn}). We also include the basic FCN used by Mask R-CNN (without guidance) for comparison. As can be seen from Table~\ref{tab:abl-ag-fcn}, AG-FCN (ours) performs the best among all the variants.


{\section{Conclusion}}

In this paper, we have presented the Fully Guided Network (FGN), a novel network to address few-shot instance segmentation. FGN can be viewed as a guided network where a support set is encoded and utilized to guide the base network, \textit{i.e.}, Mask R-CNN. Compared to previous works, FGN is characterized by introducing different guidance mechanisms into the three key components in Mask R-CNN to make full use of the guidance effect of support set. Comparative experiments on public datasets have demonstrated that FGN can outperform state-of-the-art methods. Ablation study has also been conducted to further verify the effectiveness of FGN. Despite the superiority of FGN over previous works, few-shot instance segmentation by nature is a very challenging task and there is still large room for improvement, 
especially on classification branch where more complicated features and background rejection are engaged. 
In future work, we will explore new guidance mechanisms to further boost the overall performance.  

\vspace{0.5\baselineskip}

\section*{Acknowledgement}
This work was supported by the National Natural Science Foundation of China under Grant 61703166 and Grant 61633010, the Guangdong Natural Science Foundation under Grant 2014A030312005, the Key R\&D Program of Guangdong Province under Grant 2018B030339001, the Guangzhou Science and Technology Program under Grant 201904010299,
and the Fundamental Research Funds for the Central Universities, SCUT, under Grant 2018MS72.

{\small
\bibliographystyle{ieee_fullname}
\bibliography{camera_ready}
}

\end{document}